\documentclass[letterpaper]{article} 
\usepackage{aaai2026}  
\usepackage{times}  
\usepackage{helvet}  
\usepackage{courier}  
\usepackage[hyphens]{url}  
\usepackage{graphicx} 
\usepackage{natbib}  
\usepackage{caption} 
\usepackage{algorithm}
\usepackage{algorithmic}

\usepackage{booktabs}
\usepackage{multirow}
\usepackage{amssymb}
\usepackage{amsmath}

\usepackage{newfloat}
\usepackage{listings}
\DeclareCaptionStyle{ruled}{labelfont=normalfont,labelsep=colon,strut=off} 
\floatstyle{ruled}
\newfloat{listing}{tb}{lst}{}
\floatname{listing}{Listing}
\title{GDB-Reward: From Evaluation Metrics to\\Training Rewards for Graphic Design}
\author{
    Adrienne Deganutti\textsuperscript{\rm 1,\rm 2},
    Purvanshi Mehta\textsuperscript{\rm 1},
    Simon Hadfield\textsuperscript{\rm 2},
    Andrew Gilbert\textsuperscript{\rm 2}
}
\affiliations{
    \textsuperscript{\rm 1}Lica World, 
    \textsuperscript{\rm 2}University of Surrey
}

\nocopyright
\begin{document}

\maketitle

\begin{abstract}
Text-to-image models excel at natural image synthesis but struggle with graphic design, where success depends on satisfying precise constraints on typography, layout, color, and visual communication. While prompt optimization offers an attractive alternative to expensive diffusion model fine-tuning, learning prompts for frozen image generators requires informative reward functions despite the entirely non-differentiable generation process.
Reinforcement learning does not require differentiable objectives; it requires only scalar rewards capable of ranking candidate outputs. This raises a simple question: can design evaluation metrics themselves become reinforcement learning rewards?  
Our central contribution is \textsc{GDB-Reward}, a framework that systematically transforms heterogeneous graphic design evaluation metrics into a unified reinforcement learning reward. 
Experiments demonstrate that \textsc{GDB-Reward} provides an effective optimization objective, substantially improving adherence to the design specification in perceptual quality, rendering fidelity, and spatial accuracy while keeping the image generator entirely frozen. 
More broadly, our results demonstrate that heterogeneous, non-differentiable evaluation metrics can move beyond passive benchmarking to become effective optimization objectives for reinforcement learning in domains where differentiable supervision is unavailable. 
\end{abstract}

\section{Introduction}

\begin{figure}[t!]
  \centering
  \includegraphics[width=\linewidth]{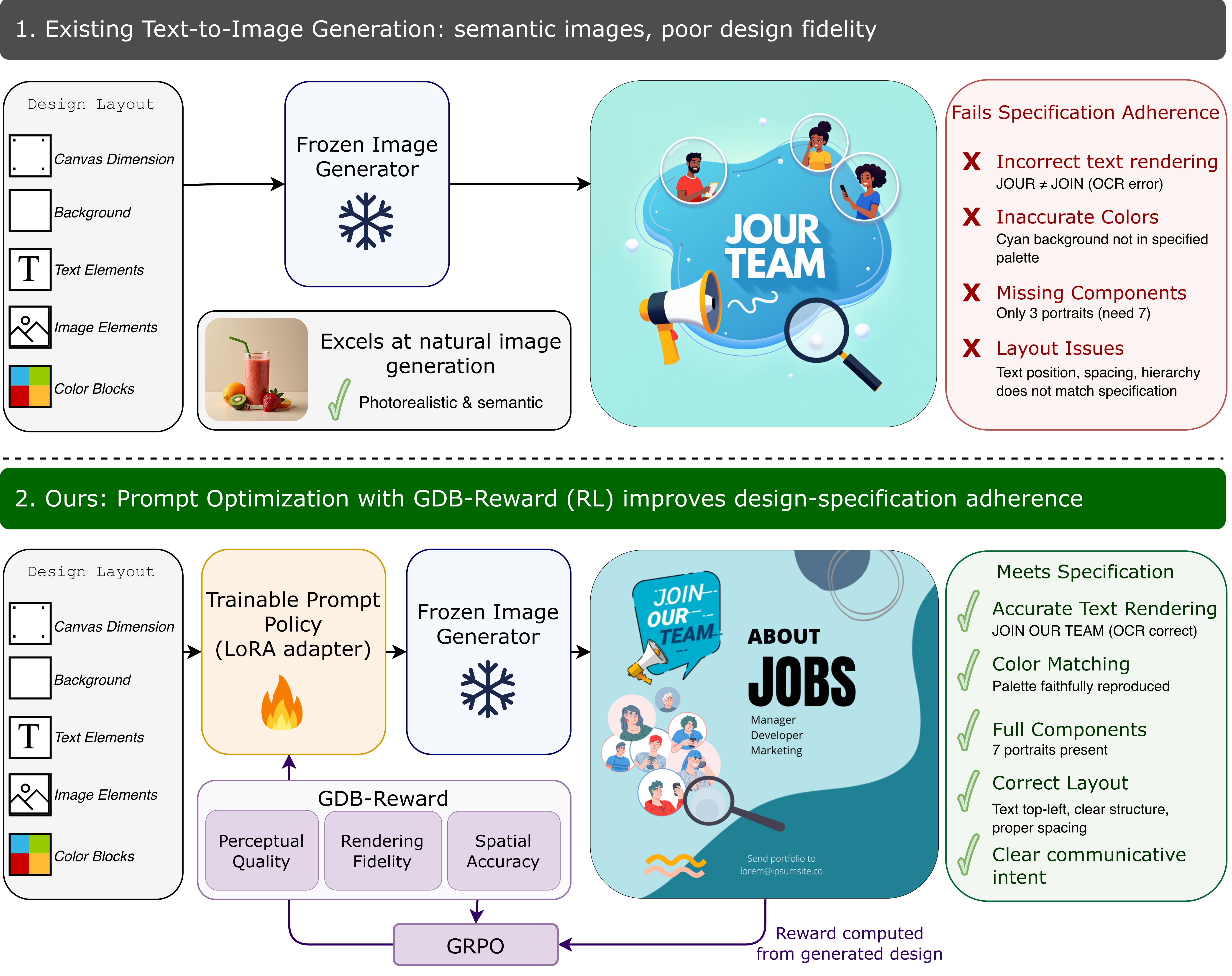}
  \caption{\textbf{Overview of the proposed framework.} Standard text-to-image models generate visually plausible images but fail to satisfy explicit design specifications. We optimize a prompt policy with reinforcement learning using \textsc{GDB-Reward}, a composite reward derived from design-specific evaluation metrics. This improves design-specification adherence while keeping the image generator frozen.}
  \label{fig:intro}
\end{figure}

Text-to-image (T2I) models generate photorealistic and stylistically rich imagery. However, graphic design requires meeting precise, structured constraints, including typography, layout, color consistency, and communicative intent. This poses a fundamentally different generation challenge. These requirements make faithful reconstruction substantially more difficult than generating natural images. A generated design may appear visually plausible yet fail to reproduce the required text, colors, or visual structure, making graphic design generation a matter of specification adherence rather than purely perceptual realism.

A natural solution could be to fine-tune the image generator itself. However, doing so requires substantial computational resources and large collections of paired, high-quality graphic design data, both of which are expensive to obtain. We instead ask a different question: \emph{can we improve specification adherence without modifying the image generator?}

The central challenge is defining an optimization objective for a completely non-differentiable generation pipeline. Unlike supervised fine-tuning, prompt optimization cannot rely on a reconstruction loss because the prompt generator never observes gradients from the image generator. Instead, learning must be driven by feedback that reflects how well the generated design satisfies the intended specification. Defining such a reward is challenging because graphic design quality is inherently multi-dimensional, requiring simultaneous assessment of typography, color, layout, semantics, and visual quality.

Recent work has begun to formalize graphic design evaluation through design-specific metrics. In particular, GraphicDesignBench~\cite{deganutti2026graphic} proposes a metric taxonomy purpose-built for design evaluation that spans spatial accuracy, typographic fidelity, structural validity, and human-aligned preference, moving beyond the perceptual quality scores (FID, CLIPScore) that dominate existing image benchmarks but fail to capture design-specific requirements. Although these metrics provide meaningful post hoc evaluation, they are not designed to serve as learning objectives because they are heterogeneous, largely non-differentiable, and combine discrete and continuous signals.

We observe that reinforcement learning does not require a differentiable objective. Instead, it requires only a scalar reward capable of ranking generated outputs. This observation changes the role of evaluation metrics: rather than designing a new reward from scratch, we transform a bank of design evaluation metrics into a unified reward for prompt optimization. This suggests a broader paradigm shift: evaluation metrics can function as optimization objectives rather than solely as passive benchmarks.

We propose \textsc{GDB-Reward}, a framework for transforming heterogeneous design evaluation metrics into a unified reinforcement learning reward. A lightweight language model rewrites structured layout metadata into natural language prompts for a frozen text-to-image model. Generated designs are scored using \textsc{GDB-Reward}, and Group Relative Policy Optimization (GRPO) updates only a LoRA adapter on the prompt generator while leaving the image generator and all reward models frozen. In this way, our approach improves adherence to the design specification without requiring expensive diffusion model fine-tuning.

Our contributions are as follows:
\begin{itemize}
    \item We propose \textsc{GDB-Reward}, a composite, non-differentiable reward that converts graphic design evaluation metrics spanning perceptual quality, rendering fidelity, and spatial accuracy into a unified optimization objective.
    \item We formulate graphic design generation as a prompt optimization problem over frozen text-to-image models, avoiding diffusion model fine-tuning through reinforcement learning.
    \item We develop a design-intent captioning model that evaluates whether generated designs preserve the underlying communicative goal, providing an independent semantic evaluation complementary to the optimization reward.
    \item We demonstrate that reinforcement learning with \textsc{GDB-Reward} substantially improves design-specification adherence, particularly for typography, color fidelity, and layout structure, while training only a lightweight LoRA adapter and leaving the image generator entirely frozen.
\end{itemize}

\section{Related Works}

\paragraph{Graphic Design Generation.} Text-to-image diffusion models excel at natural image synthesis; however, graphic design generation remains fundamentally different: success is judged not by whether an image is visually plausible but by whether it reproduces structured constraints. Multimodal systems address this by generating whole, editable designs~\citep{cheng2024graphic, jia2024colehierarchicalgenerationframework, qu2025igd}; however, these approaches train or fine-tune specialized diffusion models on paired graphic-design data, which is costly to curate and tied to a single generator. In contrast, we leave the image generator frozen and instead optimize the prompts presented to it.

\paragraph{Evaluation of graphic design.} Evaluating graphic design is a complex task because no single metric captures every axis of quality. Image evaluation has long relied on perceptual similarity measures such as SSIM~\cite{ssim_perception} and LPIPS~\cite{zhang2018perceptual}, while more recent work has introduced learned models of aesthetics and preferences. Generic text-to-image preference models such as HPSv2~\cite{wu2023human} and PickScore~\cite{kirstain2023pick} underperform in assessing design quality because they primarily focus on global image aesthetics and overlook critical dimensions of typography, layout, and structural consistency. Aesthetics$++$~\cite{kong2022aesthetics++} attempts to improve this by using a learned preference model trained on human aesthetic preferences to refine existing designs. More recently, design-specific evaluation frameworks have been proposed to better capture the unique requirements of graphic design. PosterReward~\citep{lai2026posterrewardunlockingaccurateevaluation} introduces an LLM-based evaluator that jointly assesses layout, text rendering, and aesthetic expression, while GraphicDesignBench~\cite{deganutti2026graphic} consolidates deterministic and LLM-based metrics spanning perceptual quality, text fidelity, spatial accuracy, semantic alignment, and structural validity. While prior work uses these metrics solely for evaluation, to the best of our knowledge, no prior work before \textsc{GDB-Reward} has investigated whether evaluation metrics themselves can serve as optimization objectives for reinforcement learning.

\paragraph{Prompt Optimization and Reinforcement Learning for Text-to-Image Models.} 
Rather than adapting the diffusion model itself, recent work has explored optimizing the prompts presented to frozen generators. Methods range from prompt rewriting using instruction-tuned LLMs such as Promptist~\cite{hao2023optimizing} to reinforcement learning-based approaches such as PromptLoop~\cite{lee2025plug}, demonstrating that prompt optimization can substantially improve generation quality without modifying the underlying image model. However, existing prompt optimization methods generally assume the availability of a task-specific reward or feedback signal rather than investigating how such rewards should be constructed.
The design of effective reward functions has long been recognized as a central challenge in reinforcement learning, particularly when balancing multiple competing objectives and ensuring that optimization aligns with the desired behaviour~\cite{barto2021reinforcement}. Importantly, policy-gradient methods optimize policies using scalar rewards without requiring gradients through the environment itself. Algorithms such as Proximal Policy Optimization (PPO)~\cite{schulman2017proximal} and, more recently, Group Relative Policy Optimization (GRPO)~\cite{shao2024deepseekmath} exploit only the relative ranking of candidate actions, making them naturally suited to optimization over black-box, non-differentiable objectives. Our work differs in that it focuses not on the optimization algorithm itself, but on how heterogeneous evaluation signals can be systematically transformed into reward functions for graphic design generation.

\begin{figure*}[t]
  \centering
  \includegraphics[width=\linewidth]{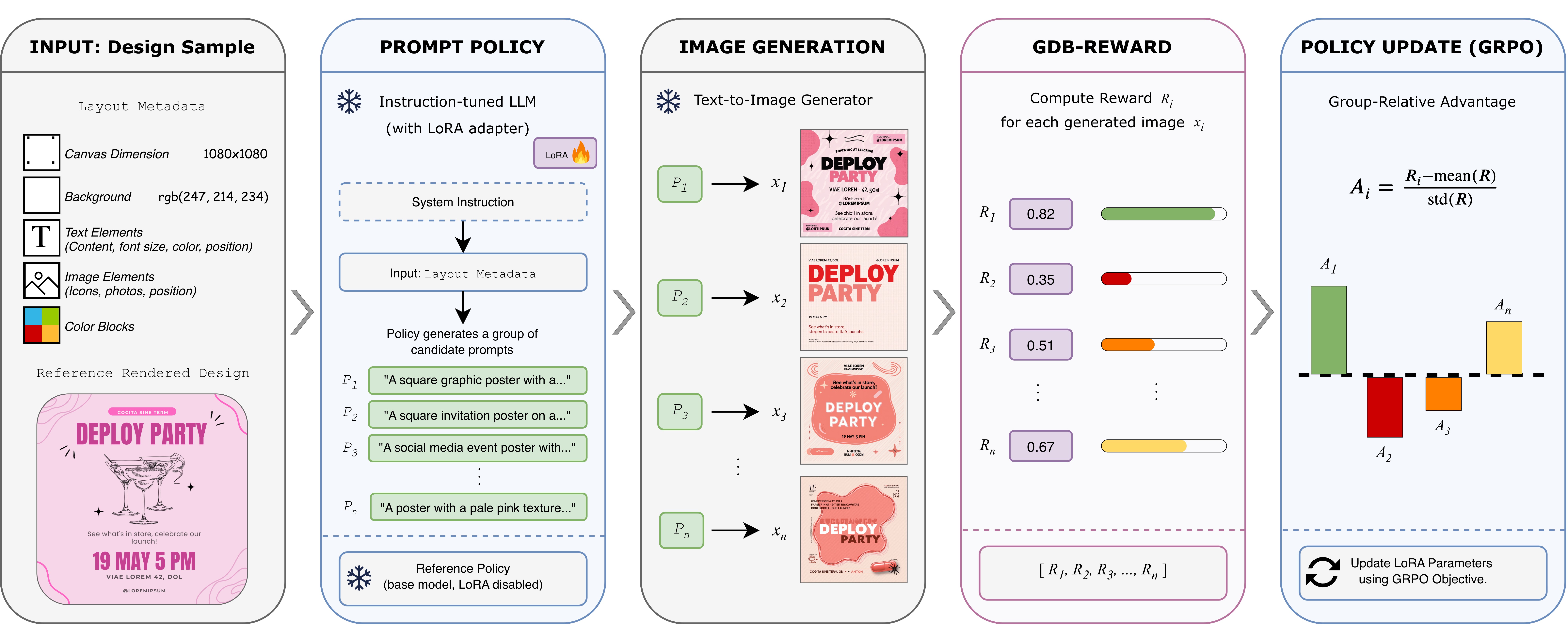}
   \caption{Overview of the proposed methodology. For each graphic design, the prompt policy maps structured layout metadata into a set of candidate image-generation prompts. A frozen text-to-image model renders each prompt into a design, which is scored by \textsc{GDB-Reward} using complementary measures of perceptual quality, rendering fidelity, and spatial accuracy. The normalized metrics are combined into a single reward that ranks candidate prompts within each group, enabling GRPO to optimize only the LoRA parameters of the prompt policy while keeping the image generator and reward function fixed.}
   \label{fig:model_overview}
\end{figure*}

Recent reinforcement learning approaches address graphic design problems but optimize different components of the pipeline. PosterReward~\cite{lai2026posterrewardunlockingaccurateevaluation} fine-tunes the diffusion model itself, LaySPA~\cite{li2026pixels} reinforces spatial reasoning over structured layout representations, and CreatiParser~\cite{chen2026creatiparser} applies GRPO to graphic design decomposition rather than image generation. In contrast, we optimize prompts using a reward systematically constructed from heterogeneous design evaluation metrics computed on the final rendered design, rather than assuming such a reward already exists.

\section{Methodology}

Our goal is to learn a prompt policy that converts structured design representations into natural-language instructions that elicit faithful reconstructions from a frozen text-to-image generator. Figure~\ref{fig:model_overview} summarizes the resulting closed loop, which is made up of four core components: the prompt generator, a frozen image generator, the design-intent model, and a composite reward, \textsc{GDB-Reward}. 

\subsection{Problem Setup}

Let a design sample consist of layout metadata $\ell$ and a rendered image $x_\ell$. The layout metadata specifies the canvas size, background, an ordered set of text elements (including their content and typographic properties), decorative image elements, and color blocks. The rendered image $x_\ell$ is an exact visualization of the layout metadata, produced using HTML and CSS visualization tools.

A policy $\pi_\theta$ (the prompt generator) maps the layout metadata $\ell$ to a textual prompt $p \sim \pi_\theta(\cdot \mid \ell)$. The generated prompt should preserve every on-design text string verbatim while describing the canvas, colors, and design components. A frozen text-to-image generator $G$ then renders an image $x = G(p)$. A reward function $R(\ell, x_\ell, x, p) \in [0,1]$ measures how faithfully the generated image $x$ matches the target design represented by $(\ell, x_\ell)$. Our objective is:
\begin{equation}
\max_{\theta}
\mathbb{E}_{\ell}
\mathbb{E}_{p \sim \pi_\theta(\cdot \mid \ell)}
\left[
R(\ell, x_\ell, x, p)
\right]
\label{eq:objective}
\end{equation}

Because neither the discrete prompt tokens nor the image generator are differentiable, we optimize Eq.~\eqref{eq:objective} using policy-gradient reinforcement learning with the scalar reward provided by \textsc{GDB-Reward}.

\subsection{Prompt Policy} 
The policy's role is to translate structured layouts into prompts that maximize \textsc{GDB-Reward}. The prompt generator is an instruction-tuned LLM equipped with a LoRA adapter~\cite{hu2022lora}. Restricting optimization to LoRA preserves the strong instruction-following capabilities of the underlying language model while providing sufficient flexibility for reward-driven prompt refinement. Given the layout metadata $\ell$, it is prompted via a fixed system instruction (see Appendix~\ref{app:method-details}) to produce a single, self-contained image-generation prompt that describes the layout, composition, palette, style, and mood while reproducing every on-design text string verbatim. As the reinforcement learning policy, the model also provides the primitives required by GRPO: sampling a group of prompts for a given layout, and computing the corresponding log-probabilities under a frozen reference policy for the KL penalty. The reference policy is initialized from the pre-trained instruction-tuned model before optimization and remains fixed throughout training.

\subsection{\textsc{GDB-Reward}} \label{method:gdb-reward}
\textsc{GDB-Reward} is a framework for transforming heterogeneous evaluation metrics into unified reinforcement learning rewards. In this work, we instantiate the framework using graphic design evaluation metrics. We combine complementary signals covering three design-native dimensions drawn from the GraphicDesignBench taxonomy~\citep{deganutti2026graphic}: perceptual quality, rendering fidelity, and spatial accuracy. Designing a reward for reinforcement learning requires satisfying three properties: (i) heterogeneous evaluation signals must be brought onto a common scale; (ii) the reward must remain numerically stable across samples; and (iii) it must preserve the relative ordering of candidate generations. Since these metrics produce heterogeneous outputs (e.g., perceptual distances, OCR accuracy, cosine similarity), each score is mapped to a normalized reward in $[0,1]$, where higher values indicate better performance, and the scores are then combined into a single scalar reward. The selected metrics satisfy two design principles. First, they collectively capture complementary aspects of design quality that a single metric cannot measure. Second, each produces a reliable scalar score that can be normalized and combined into an effective optimization signal for reinforcement learning.

Formally, let $\mathcal{M}$ denote the set of metrics with non-negative weights $\{w_k\}_{k\in\mathcal{M}}$. Each metric is represented by a measurement function $\mu_k : (\ell, x_\ell, x, p) \rightarrow \mathbb{R}$ which computes the raw score produced by metric $k$. The resulting measurement is:
\begin{equation}
   m_k=\mu_k(\ell,x_\ell,x,p) 
\end{equation}

where $\ell$ is the layout metadata, $x_\ell$ is the reference rendering, $x$ is the generated image, and $p$ is the generated prompt. Because different metrics produce values in different units (e.g., LPIPS distances, OCR accuracy, or color differences), each measurement is converted into a normalized reward:
\begin{equation}
    r_k=\phi_k(m_k)\in[0,1]
\label{eq:composite-reward}
\end{equation}

where larger values always indicate better design quality. Keeping the raw measurements separate from their normalized rewards allows us to report standard evaluation metrics while presenting the reinforcement-learning algorithm with a consistent reward scale.

We organize the metric bank according to the following design objectives, summarized in Table~\ref{tab:metrics}.

\paragraph{Perceptual Quality.} Perceptual quality metrics measure the visual similarity between the generated image and the intended design. We use three reference-based measures: DreamSim~\cite{fu2023dreamsim}, LPIPS~\cite{zhang2018perceptual}, and SSIM~\cite{ssim_perception}. DreamSim and LPIPS produce perceptual distances that are converted into rewards using: 
\begin{equation}
    \phi(m)=e^{-m/\tau} \quad\text{where}\quad \tau=20
\label{eq:exp-normalization}
\end{equation}

\noindent while SSIM is rescaled from $[-1,1]$ to $[0,1]$ by:
\begin{equation}
    \phi(m)=\frac{m+1}{2}
\label{eq:normalize}
\end{equation}

To provide reference-free quality signals, we additionally include ImageReward~\cite{xu2023imagereward} and NIMA~\cite{Talebi_2018}. ImageReward scores are passed through a logistic sigmoid, while NIMA's predicted aesthetic score is linearly mapped from $[1,10]$ to $[0,1]$.

\paragraph{Rendering fidelity.} Graphic designs often depend on accurate typography and color reproduction. We therefore evaluate three complementary properties. Color fidelity measures the average CIEDE2000~\cite{sharma2005ciede2000} distance ($\Delta E_{00}$) between the colors specified in the layout metadata and the nearest dominant color extracted from the generated image; we then convert it into a reward using Eq.~\eqref{eq:exp-normalization}. Typography is evaluated using OCR read-back accuracy, which compares each expected text string with the OCR output bounded in $[0,1]$.

\paragraph{Spatial Accuracy.}\label{method:spatial-accuracy} Spatial accuracy evaluates whether the correct number of components are rendered. We detect text and image regions using a pre-trained YOLO-OBB detector from the GraphicDesignBench repository~\citep{deganutti2026graphic} that was trained on graphic design layouts. We compare the detections with both the reference render and the layout metadata. Component count penalizes discrepancies in the number of detected elements: $\phi(m)=\max(0,1-m)$. 

\begin{table}[h]
\centering
\small
\begin{tabular}{@{}lll@{}}
    \toprule
    Category & Metric & Evaluation Signal \\
    \midrule
    \multirow{5}{*}{Perceptual}
    & DreamSim    & perceptual distance \\
    & LPIPS       & perceptual distance \\
    & SSIM        & structural similarity \\
    & ImageReward & learned human preference \\
    & NIMA        & aesthetic preference \\
    \midrule
    \multirow{2}{*}{Rendering}
    & $\Delta E_{00}$  & color distance \\
    & OCR         & text reconstruction \\
    \midrule
    \multirow{1}{*}{Spatial}
    & YOLO Count       & component count error \\
    \bottomrule
\end{tabular}
\caption{The metric bank used by \textsc{GDB-Reward}.}
\label{tab:metrics}
\end{table}

\begin{table*}[t]
\centering
\small
\setlength{\tabcolsep}{5pt}
\begin{tabular}{@{}lcccccccccc@{}}
\toprule
& & \multicolumn{5}{c}{Perceptual quality} &
\multicolumn{2}{c}{Rendering fidelity} & \multicolumn{2}{c}{Spatial accuracy} \\
\cmidrule(lr){3-7}\cmidrule(lr){8-9}\cmidrule(lr){10-11}
& & \multicolumn{3}{c}{Reference-based} & \multicolumn{2}{c}{Reference-free} & & & & \\
\cmidrule(lr){3-5}\cmidrule(lr){6-7}
Model & Full & DS & LP & SS & IR & NM & $\Delta E_{00}$ & OCR & C$_g$ & C$_m$ \\
\midrule
Base (Qwen3.5-9B $+$ FLUX.2-dev)
 & 0.560 & 0.543 & \textbf{0.314} & 0.744 & 0.366 & \textbf{0.418} & 0.462 & \textbf{0.870} & 0.591 & 0.482 \\
$+$\textsc{GDB-Reward} (trained)
 & \textbf{0.647} & \textbf{0.565} & 0.296 & 0.735 & \textbf{0.571} & 0.415 & \textbf{0.578} & 0.869 & \textbf{0.731} & \textbf{0.675} \\
\midrule
Base (Qwen3.5-9B $+$ FLUX.1-dev)
 & 0.398 & 0.437 & 0.336 & 0.768 & 0.337 & 0.387 & 0.378 & 0.446 & 0.353 & 0.307 \\
$+$\textsc{GDB-Reward} (trained)
 & \textbf{0.559} & \textbf{0.513} & 0.272 & 0.730 & \textbf{0.545} & \textbf{0.427} & \textbf{0.504} & \textbf{0.643} & \textbf{0.601} & \textbf{0.612} \\
\bottomrule
\end{tabular}
\caption{Per-metric performance on the LICA test split. Entries are the mean normalized metric score $\phi_k \in [0,1]$ (higher is better). The composite Full score is upper-bounded at $0.758$ rather than $1$, since reference-based perceptual metrics cannot saturate even for the ground-truth render; trained scores should be read against this ceiling. \textbf{Perceptual:} DS=DreamSim, LP=LPIPS, SS=SSIM, IR=ImageReward, NM=NIMA; \textbf{Rendering:} $\Delta E_{00}$=CIEDE2000 color distance, OCR=OCR accuracy; \textbf{Spatial:} C$_g$/C$_m$=YOLO component count vs.\ ground-truth render / layout metadata.}
\label{tab:main}
\end{table*}

\paragraph{Composite Reward.} Not every metric applies to every design; OCR metrics, for example, cannot be computed when no text is present. Let $A(\ell,x)\subseteq\mathcal{M}$ denote the subset of applicable metrics used to produce measurements for the current sample. The final reward is computed as the weighted arithmetic mean over these metrics. We adopt a weighted arithmetic mean rather than multiplicative or minimum-based aggregation because individual metrics capture complementary, not mandatory, properties of a successful design. Product-based aggregation would excessively penalize isolated weaknesses, while minimum-based aggregation would cause optimization to be dominated by a single poorly performing metric. The weighted average instead provides smooth trade-offs while preserving the contribution of all metrics.
\begin{equation}
    R(\ell,x_\ell,x,p)=
    \frac{
    \sum_{k\in A(\ell,x)}
    w_k\,
    \phi_k(\mu_k(\ell,x_\ell,x,p))
    }{
    \sum_{k\in A(\ell,x)}
    w_k
    }
\label{eq:gdb-reward}
\end{equation}


Finally, \textsc{GDB-Reward} is designed specifically for Group Relative Policy Optimization. Because GRPO standardizes rewards within each sampled group, only the relative ranking of candidate prompts affects the policy update. Consequently, \textsc{GDB-Reward} is designed to reliably distinguish better reconstructions from worse ones, rather than to produce perfectly calibrated absolute scores.

\subsection{Policy Optimization with GRPO}
\label{method:grpo}

We optimize the prompt generator using GRPO, treating it as the policy and \textsc{GDB-Reward} as a black-box reward function. Since the generated prompt consists of discrete tokens and the text-to-image generator is non-differentiable, gradients cannot be propagated through the rendering process. Reinforcement learning instead allows the prompt generator to improve by maximizing the scalar reward assigned to the generated image.

For each layout $\ell$, the policy samples a group of $G$ candidate prompts:
\begin{equation}
    \{p_i\}_{i=1}^{G} \sim \pi_\theta(\cdot \mid \ell)
\end{equation}

where $\pi_\theta$ denotes the prompt generator. The frozen text-to-image model renders each prompt to produce an image $x_i = G(p_i)$, which is then evaluated using \textsc{GDB-Reward} from Eq.~\eqref{eq:gdb-reward}. GRPO computes a group-relative advantage by standardizing rewards within the sampled group:
\begin{equation}
    A_i =
    \frac{
    R_i-\mathrm{mean}(\{R_j\})
    }{
    \mathrm{std}(\{R_j\})+\epsilon
    }
\end{equation}

This encourages prompts that outperform the group average while discouraging poorer candidates. The prompt generator is updated using the GRPO policy-gradient objective with a KL regularization term that constrains the policy to remain close to the frozen base language model. Throughout training, only the LoRA adapter of the prompt generator is updated. Both the image generator and \textsc{GDB-Reward} remain frozen and operate entirely in inference mode.

\begin{figure*}[t]
  \centering
  \includegraphics[width=\linewidth]{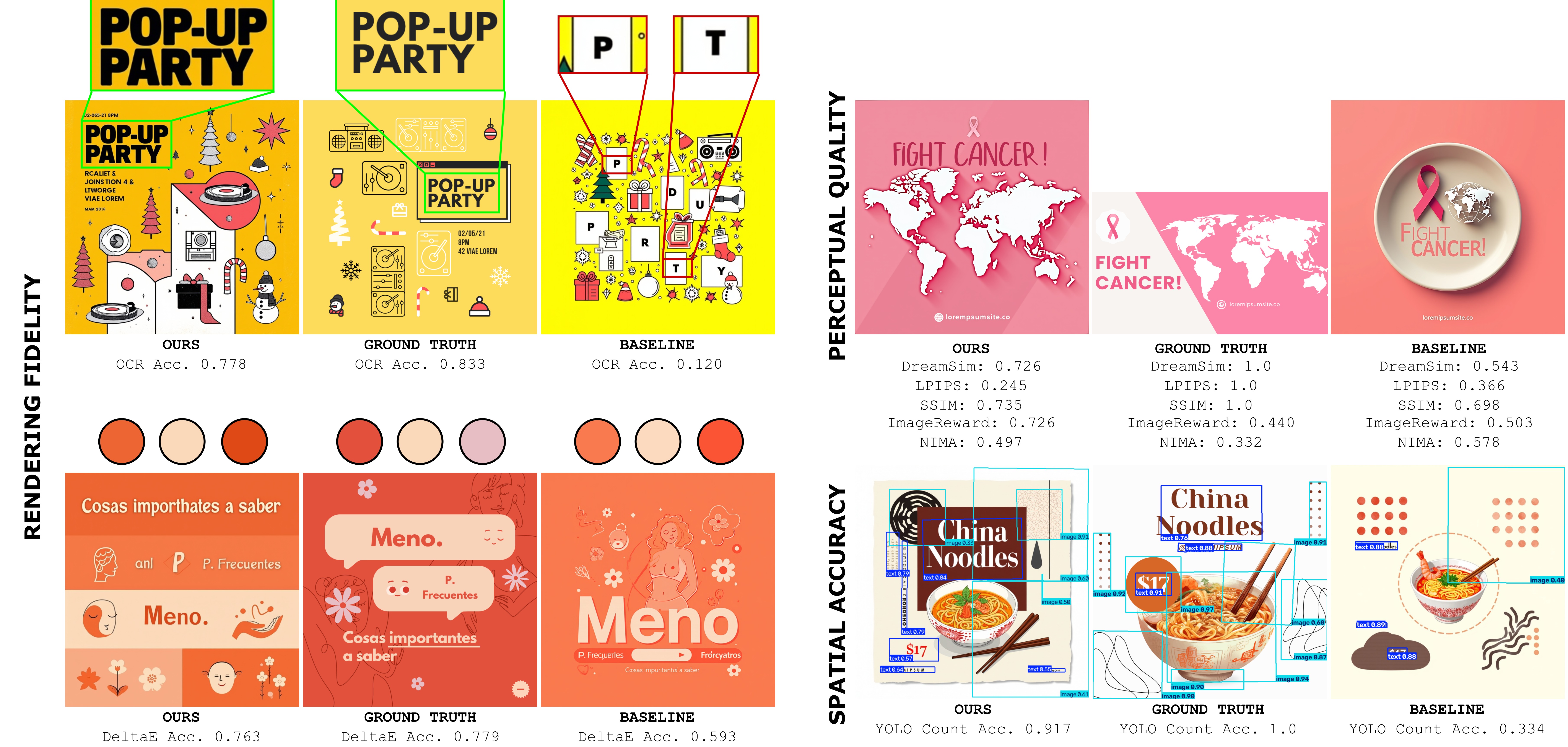}
   \caption{\textbf{Qualitative results.} Comparison of designs generated by our method and the baseline, both using Qwen3.5-9B and FLUX.1-dev as backbones. The examples highlight improvements across the four evaluation dimensions: rendering fidelity, perceptual quality, and spatial accuracy.}
   \label{fig:qualitative}
\end{figure*}

\section{Experiments}

\subsection{Implementation Details}

The prompt policy is Qwen3.5-9B~\cite{qwen3.5} fine-tuned with LoRA adapters ($r=16$, $\alpha=32$), optimizing $29.1$M parameters while keeping the backbone frozen. We train with GRPO using frozen FLUX.1-dev~\cite{flux2024} and FLUX.2-dev~\cite{flux-2-2025} generators and our proposed \textsc{GDB-Reward}. Further implementation details are provided in Appendix~\ref{app:exp-details}.

\subsection{Datasets}

Experiments are conducted on the LICA graphic design benchmark \cite{deganutti2026graphic}. Each example contains structured layout metadata, a corresponding reference image, and a concise textual description of the intended design objective (examples shown in Appendix~\ref{app:dataset}). We train on 911 layouts and evaluate on a held-out test split of 100 layouts.

\subsection{\textsc{GDB-Reward}: Quantitative Analysis}

Table~\ref{tab:main} evaluates whether \textsc{GDB-Reward} functions as an effective optimization objective. We compare the frozen prompt policy with the same policy after reinforcement learning using \textsc{GDB-Reward}, holding the image generator fixed and repeating the experiment across two generators of differing capability. Improvements therefore isolate the reward's ability to guide optimization 
across diffusion models with a range of capabilities. It is worth noting that this evaluation is performed using the same metrics that comprise our training loss. The evaluations in later sections will provide independent verifications of the performance.

\textsc{GDB-Reward} increases the composite score under both generators, confirming that its optimization signal is 
consistent across generators of different sizes.
The gains are concentrated on the metrics that directly test specification adherence: OCR accuracy, color fidelity, and component-count agreement. Two properties show these gains are principled rather than incidental. First, the reward targets each backbone's failure mode: on FLUX.1, where text rendering is weak, OCR jumps $0.446\!\to\!0.643$, whereas on FLUX.2 OCR already sits at ceiling. Second, training pulls the two generators toward a common specification-satisfying operating point: the per-metric gap between them shrinks after training, and the trained FLUX.1 policy ($0.559$) recovers essentially the full quality of the untrained FLUX.2 baseline ($0.560$) through prompting alone. The perceptual metrics further validate this objective. Reference-free quality improves consistently, while reference-based similarity is mixed, with DreamSim increasing but LPIPS and SSIM decreasing slightly. This is expected because a design brief allows multiple valid realizations; optimizing for the specification preserves semantic fidelity while allowing some variation in typography, spacing, and placement.

\subsection{\textsc{GDB-Reward}: Qualitative Analysis}
Figure~\ref{fig:qualitative} provides qualitative examples illustrating the behavior observed quantitatively. Rather than consistently producing more aesthetically pleasing images, the optimized prompt policy generates designs that more closely satisfy the intended design specification. Improvements are most evident in three areas. First, generated text is reproduced more accurately and legibly, consistent with the large increase in OCR accuracy. Second, color palettes more closely match the specified design colors, reflecting the improvement in color fidelity. Third, layouts contain a more appropriate number of visual components and better preserve the intended composition, matching the gains in the spatial evaluation metrics. These examples mirror the quantitative evaluation and suggest that reinforcement learning improves specification adherence. Additional qualitative results are given in Appendix~\ref{app:qualitative}.

\subsection{Ablations}
\paragraph{Task groups.} To evaluate the contribution of each reward category, we retrain the prompt policy while removing one of the three groups of reward terms: perceptual quality, rendering fidelity, or spatial accuracy. Table~\ref{tab:ablation-categories} reports the resulting composite scores. Removing any reward group degrades overall performance, indicating that all three contribute to the final policy. The largest reduction occurs when rendering fidelity is removed, suggesting that fine-grained content guidance provides the strongest optimization signal. Nevertheless, combining all three categories consistently achieves the highest overall reward, demonstrating that the different objectives are complementary.

\begin{table}[h]
\centering
\setlength{\tabcolsep}{5pt}
\begin{tabular}{@{}lc@{}}
    \toprule
    Training reward & Composite Score \\
    \midrule
    \midrule
    Full \textsc{GDB-Reward} & \textbf{0.530} \\
    \midrule
    Perceptual quality $+$ Rendering fidelity & 0.523 \\
    Rendering fidelity $+$ Spatial accuracy & 0.498 \\
    Spatial accuracy $+$ Perceptual quality  & 0.491 \\
    \bottomrule
\end{tabular}
\caption{Ablation of the three \textsc{GDB-Reward} task groups, trained on a 50-sample subset of the training dataset, and evaluated on the full evaluation split. Each row is the Qwen3.5-9B prompt generator trained with one group removed from the reward, paired with FLUX.1-dev.}
\label{tab:ablation-categories}
\end{table}

\begin{table*}[t]
\centering
\small
\setlength{\tabcolsep}{5pt}
\begin{tabular}{@{}llcccccccccc@{}}
\toprule
 & & & \multicolumn{5}{c}{Perceptual quality} &
\multicolumn{2}{c}{Rendering fidelity} & \multicolumn{2}{c}{Spatial accuracy} \\
\cmidrule(lr){4-8}\cmidrule(lr){9-10}\cmidrule(lr){11-12}
& & Full & DS & LP & SS & IR & NM & $\Delta E_{00}$ & OCR & C$_g$ & C$_m$ \\
\midrule
Equal per concept & \textit{Weight}
 & -- & \textit{0.050} & \textit{0.050} & \textit{0.050} & \textit{0.050} & \textit{0.050} & \textit{0.125} & \textit{0.125} & \textit{0.125} & \textit{0.125} \\
(default) & Score
 & \textbf{0.530} & \textbf{0.514} & 0.300 & 0.738 & \textbf{0.514} & 0.426 & 0.468 & \textbf{0.709} & \textbf{0.547} & \textbf{0.464} \\
\midrule
Rendering and & \textit{Weight}
 & -- & \textit{0.030} & \textit{0.020} & \textit{0.005} & \textit{0.040} & \textit{0.005} & \textit{0.160} & \textit{0.300} & \textit{0.220} & \textit{0.220} \\
spatial-heavy & Score 
 & 0.508 & 0.495 & \textbf{0.316} & \textbf{0.743} & 0.431 & \textbf{0.433} & \textbf{0.491} & 0.648 & 0.507 & 0.437 \\
\bottomrule
\end{tabular}
\caption{Reward-reweighting ablation. Both rows train the Qwen3.5-9B $+$ FLUX.1-dev prompt policy with \textsc{GDB-Reward} under identical GRPO settings and differ only in the training reward weights. \emph{Equal per concept} is our default (Table~\ref{tab:metrics}).}
\label{tab:ablation-reweight}
\end{table*}

\paragraph{Reward-reweighting.} A separate design question is whether the relative weighting of metrics within \textsc{GDB-Reward} affects performance. Our default weighting assigns each of the three task groups equal total importance, split evenly across its metrics. Since GRPO normalizes rewards within each sampled group, optimization is primarily driven by the relative ordering of generated samples rather than the absolute scale of individual reward terms. We stress-test this by retraining with a deliberately skewed composite that shifts roughly $90\%$ of the reward mass onto the higher-variance, more discriminating rendering-fidelity (OCR, $\Delta E_{00}$) and spatial (YOLO component-count) metrics. Table~\ref{tab:ablation-reweight} demonstrates that despite devoting far more training signal to rendering fidelity and spatial accuracy, the reweighted policy is no better on the very axes it emphasizes. This corroborates the principle that within-group standardization already rescales each metric by its group variance, making hand-tuned per-metric coefficients largely redundant and occasionally harmful, which justifies the simple equal-per-concept default.

\subsection{\textsc{GDB-Reward} vs. Generator Fine-tuning}

Our central design decision is to optimize the prompt policy while keeping the image generator frozen. We compare this approach against supervised fine-tuning (SFT) of the image generator itself. The SFT method attaches a LoRA adapter to the image generator and trains the denoising network using a standard flow-matching objective to reconstruct the reference render directly from the layout metadata. In contrast, our approach leaves the generator unchanged and optimizes only the language model that produces the prompt. Table~\ref{tab:grpo-vs-sft} compares both approaches. SFT lands below the frozen baseline on both backbones. We attribute this to an objective-and-conditioning mismatch. SFT conditions the denoiser on serialized layout metadata which is far out of distribution for a generator whose text encoder expects natural-language prompts. This pulls the model off its strong pretrained prior toward exact reconstruction of one target, trading away the perceptual and aesthetic quality the composite score rewards.

\begin{table}[h]
\centering
\small
\setlength{\tabcolsep}{5pt}
\begin{tabular}{lcccc}
    \toprule
    \multirow{2}{*}{Method} & \multicolumn{2}{c}{Trainable Params.} & \multicolumn{2}{c}{Composite Score} \\
    \cmidrule(lr){2-3} \cmidrule(lr){4-5}
     & FL.1 & FL.2 & FL.1 & FL.2 \\
    \midrule
    \midrule
        SFT & 46.7M & 83.4M & 0.342 & 0.518 \\
    \midrule
        Baseline & - & - & 0.398 & 0.560 \\
    \midrule
        \textsc{GDB-Reward} & 29.1M & 29.1M & \textbf{0.559} & \textbf{0.647} \\
    \bottomrule
\end{tabular}
\caption{Comparison of supervised generator fine-tuning (SFT) and \textsc{GDB-Reward} prompt optimization. FL.1 and FL.2 denote FLUX.1-dev and FLUX.2-dev.}
\label{tab:grpo-vs-sft}
\end{table}

\vspace{-0.4cm}

\subsection{Design-Intent Evaluation}
\label{sec:design-intent}

While \textsc{GDB-Reward} evaluates low-level aspects of graphic design quality, it does not explicitly measure whether a generated design preserves the original communicative goal. We therefore perform a complementary evaluation based on design intent. Given a generated design, a vision--language model predicts a concise textual description of its intended deliverable, message, target audience, and overall tone. This predicted description is compared against the ground-truth design intent provided by the LICA dataset. Details of the evaluator and its training are provided in Appendix~\ref{appendix:design-intent}. We evaluate design-intent preservation as a text generation task using both semantic and lexical similarity metrics. Our primary metric is G-Eval~\cite{liu2023gevalnlgevaluationusing}, implemented with DeepEval~\cite{ip2026deepeval} using GPT-5.5~\cite{openai2026gpt55}, which measures whether the predicted description captures the same underlying design objective as the reference. We additionally report BLEU-1~\citep{papineni2002bleu}, METEOR~\citep{banerjee2005meteor}, ROUGE-L~\citep{lin2004rouge}, and CIDEr~\citep{vedantam2015cider}, which quantify lexical agreement between the predicted and reference intents.

Table~\ref{tab:intent-recon} shows that design-intent preservation is primarily limited by the image generator rather than the prompt policy: upgrading the frozen backbone from FLUX.1-dev to FLUX.2-dev improves G-Eval from $52.5$ to $65.0$. Across all metrics, \textsc{GDB-Reward} predominantly preserves or slightly improves design intent, demonstrating that the reward improves measurable design fidelity without sacrificing the original communicative goal.

\begin{table}[h]
\centering
\small
\setlength{\tabcolsep}{4pt}
\begin{tabular}{@{}lccccc@{}}
\toprule
    Prompt policy & B1 $\uparrow$ & M $\uparrow$ & R-L $\uparrow$ & Cr $\uparrow$ & G-Eval $\uparrow$ \\
\midrule
    \multicolumn{6}{l}{\textit{FLUX.1-dev}} \\
\midrule
    Base & 41.28 & 16.48 & 31.75 & 50.86 & 52.50 \\
    $+$\textsc{GDB-Reward} & \textbf{42.19} & \textbf{17.09} & \textbf{32.14} & \textbf{50.92} & \textbf{52.70} \\
\midrule
    \multicolumn{6}{l}{\textit{FLUX.2-dev}} \\
\midrule
    Base & 44.18 & \textbf{19.27} & 35.72 & 65.72 & \textbf{65.00} \\
    $+$\textsc{GDB-Reward} & \textbf{45.29} & 18.97 & \textbf{35.87} & \textbf{67.66} & 61.60 \\
\bottomrule
\end{tabular}
\caption{Design-intent preservation of the generated reconstructions on the LICA test split, scored by the fine-tuned Qwen3.5-9B captioner against the ground-truth intent.}
\label{tab:intent-recon}
\end{table}

\vspace{-0.5cm}

\section{Conclusion}

We presented \textsc{GDB-Reward}, a framework that transforms heterogeneous graphic design evaluation metrics into a unified reinforcement learning reward for prompt optimization. By optimizing only a lightweight prompt policy while keeping the text-to-image generator frozen, \textsc{GDB-Reward} improves adherence to graphic design specifications without diffusion model fine-tuning. More broadly, our results suggest that evaluation metrics can serve not only as benchmarks but also as effective optimization objectives for reinforcement learning in non-differentiable generation tasks.

\bibliography{aaai2026}

\clearpage
\appendix

\section{Additional Methodology Details} \label{app:method-details}

\paragraph{Prompt Policy} The prompt generator used as our policy is an instruction-tuned LLM equipped with a LoRA adapter. Given the layout metadata, the model is prompted via the fixed system instruction shown in Figure~\ref{fig:app:prompt-generator-prompt}.

\begin{figure}[h]
  \centering
  \fbox{
  \begin{tabular}{p{0.9\columnwidth}}
   You are an expert prompt engineer for a text-to-image model. You are given the structured metadata of a finished graphic design (its canvas, colours, text content and decorative elements). Rewrite this metadata into ONE vivid, self-contained natural-language image-generation prompt that would recreate the design. \\
   Rules: \\
    - Write a single flowing paragraph, no lists, no preamble, no quotes. \\
    - Describe layout, composition, colour palette, style and mood concretely. \\
    - Include every piece of on-design text verbatim, in quotes, with where it sits. \\
    - Do not mention JSON, metadata, pixels, or coordinates. \\
    - Output only the prompt.
  \end{tabular}}
  \caption{System instruction used to condition the prompt policy. The instruction guides the LLM to convert structured graphic design metadata into a single natural-language prompt for the frozen text-to-image generator.}
  \label{fig:app:prompt-generator-prompt}
\end{figure}

\section{Experiments Implementation Details}
\label{app:exp-details}

This section provides the implementation details required to reproduce our experiments, including the prompt policy, frozen image generator, \textsc{GDB-Reward} configuration, and reinforcement learning hyperparameters.

\paragraph{Prompt policy.} The prompt policy is Qwen3.5-9B operating in \texttt{bfloat16}, fine-tuned with LoRA adapters ($r=16$, $\alpha=32$) inserted into the attention and MLP projection layers while keeping the backbone frozen. In total, $29.1$M parameters ($\sim0.2\%$ of the model) are optimized. During training, prompts are generated from the layout metadata using nucleus sampling (temperature $1.0$, top-$p=0.95$) with a maximum length of $320$ tokens.

\paragraph{Image generator.} We report results on both FLUX.1-dev and FLUX.2-dev as the frozen image generator. Images are generated using $20$ diffusion steps with guidance scale $3.5$ and sequence length $512$, producing $768\times768$ images during RL training and $1024\times1024$ images for evaluation.

\paragraph{\textsc{GDB-Reward} configuration.} Our proposed \textsc{GDB-Reward} uses six dominant palette colors ($K=6$) and color sharpness parameter $\tau=20$. For layout analysis, we use a fine-tuned YOLO11x-OBB detector taken from the GraphicDesignBench~\cite{deganutti2026graphic} repository\footnote{\url{https://github.com/purvanshi/lica-bench/tree/main/scripts/motion_gen_eval}}. All reward models operate in inference mode only.

\paragraph{RL hyperparameters.} We optimize the prompt policy using GRPO for $1368$ optimization steps (three epochs) when using FLUX.1-dev as the backbone, and $912$ optimization steps (two epochs) when using FLUX.2-dev. We set the GRPO group size to $12$ sampled prompts across two layouts per update. The learning rate is $3\times10^{-5}$, the KL coefficient is $\beta=0.02$, gradients are clipped to $1.0$, and the advantage standard deviation is lower bounded by $10^{-4}$. Following prior work, common random numbers (shared diffusion seeds) are used within each group to reduce sampling variance. All experiments use a fixed random seed of $0$.

\paragraph{Computing infrastructure.}
All experiments were conducted on a compute node equipped with eight NVIDIA H100 SXM5 GPUs (80\,GB memory each), 168 CPU cores, 907\,GB system memory, and 12.2\,TB of local storage, running Ubuntu~24.04 LTS. The implementation was built using PyTorch, Hugging Face Transformers, TRL for GRPO optimization, PEFT for LoRA fine-tuning, and Diffusers for image generation. The experiments were performed with CUDA~12.0.

\section{Dataset Examples}
\label{app:dataset}

\begin{figure*}[t!]
  \centering
  \includegraphics[width=\linewidth]{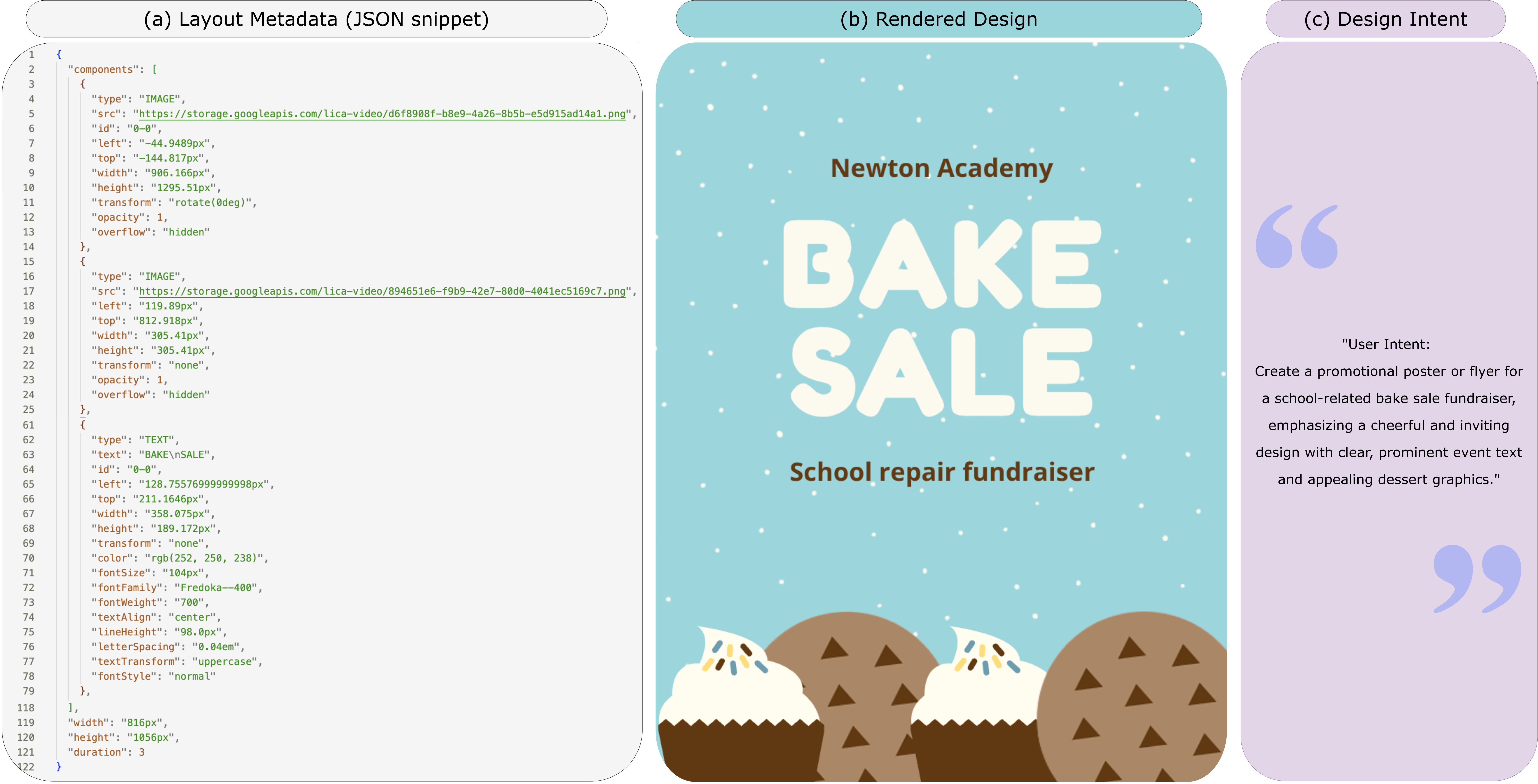}
  \caption{\textbf{Example from the LICA GraphicDesignBench dataset.} Each sample consists of (a) serialized layout metadata describing the design elements and their attributes, (b) the corresponding rendered graphic design, and (c) a natural-language design intent used to guide generation. Metadata has been truncated for readability.}
  \label{fig:data-example}
\end{figure*}

We provide a full example from the LICA GraphicDesignBench dataset~\cite{deganutti2026graphic}. Figure~\ref{fig:data-example} illustrates the dataset structure, showing the design intent description, rendered image, and corresponding layout metadata for each example.

\section{Design-Intent Evaluation Model}
\label{appendix:design-intent}

To evaluate whether generated designs preserve their intended communicative goal, we train a dedicated vision--language model that predicts a textual description of the design intent from a rendered graphic design. Each description summarizes the intended deliverable, message, target audience, and desired emotional tone. The model is used exclusively for evaluation and is never incorporated into the reinforcement-learning reward.

\begin{table}[h]
  \centering
  \small
  \renewcommand{\arraystretch}{1.2}  
  \begin{tabular}{c|cccc|c}  
    \toprule
    \multirow{2}{*}{Model} & \multirow{2}{*}{B1 $\uparrow$} & \multirow{2}{*}{M $\uparrow$} & \multirow{2}{*}{R-L $\uparrow$} & \multirow{2}{*}{Cr $\uparrow$} & G-Eval $\uparrow$ \\
     & & & & & (gpt-5.5) \\
    \midrule
    \midrule
    \multicolumn{6}{l}{\textit{Base model}} \\
    \midrule
    Qwen-3.5 & \textbf{19.60} & 13.57 & 17.39 & \textbf{4.69} & \textbf{55.80} \\
    Llama-4 & 17.96 & \textbf{14.89} & \textbf{19.35} & 3.98 & 53.00 \\
    DeepSeek & 6.91 & 10.97 & 09.61 & 0.00 & 19.30 \\
    \midrule
    \midrule
    \multicolumn{6}{l}{\textit{Fine-tuned model}} \\
    \midrule
    Qwen-3.5 & 48.84 & 21.90 & \textbf{39.97} & \textbf{87.25} & \textbf{72.90} \\
    Llama-4 & \textbf{49.34} & \textbf{21.99} & 39.42 & 85.73 & 72.60 \\
    DeepSeek & 47.92 & 20.74 & 37.13 & 73.69 & 67.50 \\
    \bottomrule
  \end{tabular}
  \caption{We finetune four models: Qwen3.5-9B, Llama-4-Scout-17B-16E-Instruct, and DeepSeek-VL-7B-Chat on our design intent captioning task and report both semantic and exact match metrics on the test split.}
  \label{tab:app-design-intent}
\end{table}

We fine-tune pretrained vision--language models using LoRA on paired design renders and intent descriptions from the LICA training split. Training follows a standard autoregressive objective, where the model learns to generate the corresponding design-intent description given the rendered design image. During evaluation, the predicted intent $\hat{y}=f_{\phi}(x)$ is compared with the ground-truth intent $y_\ell$, where $x$ is the generated image.

\medskip
To select the evaluation model, we first compare several candidate vision--language backbones on the held-out LICA test split. Table~\ref{tab:app-design-intent} reports both the performance of the original pretrained models and their fine-tuned counterparts. We evaluate using G-Eval~\cite{liu2023gevalnlgevaluationusing} as the primary semantic metric. We provide the prompt and evaluation rubric used by G-Eval to assess the design-intent in Figure~\ref{fig:app:g-eval-prompt} and Table~\ref{tab:app:g-eval-rubric}. Furthermore, we evaluate the VLM using BLEU-1~\citep{papineni2002bleu}, METEOR~\citep{banerjee2005meteor}, ROUGE-L~\citep{lin2004rouge}, and CIDEr~\citep{vedantam2015cider}. Fine-tuning substantially improves performance for all candidate models. Although Qwen3.5 and Llama-4 obtain comparable lexical scores, the fine-tuned Qwen3.5 model achieves the highest G-Eval score and is therefore selected as the design-intent evaluator used throughout our experiments.
 
\newpage
\begin{figure}[h!]
  \centering
  \small
  \fbox{
  \begin{tabular}{p{0.9\columnwidth}}
   Note the FORM of 'expected output': it is a single, concise design-intent statement (typically one sentence, roughly 20-40 words) that names the deliverable/artifact, its message or purpose, the target audience, and the intended feeling or tone. Identify those core elements (deliverable, message/purpose, audience, feeling/tone) in 'expected output' and check whether 'actual output' states the same core goal. Reward semantic agreement even when the wording differs. Judge concision and form: 'actual output' should be about as brief as 'expected output' and read as a single-paragraph intent statement. HEAVILY penalize output that is markedly longer or more verbose than 'expected output' -- multiple sentences or paragraphs, exhaustive visual description, enumerating colours/objects/positions, step-by-step or list formatting, or reading like an image caption rather than a concise statement of intent -- EVEN IF the extra content is accurate. Heavily penalize contradictions (wrong artifact type, wrong audience, opposite mood) and omission of a core element present in 'expected output'. Do NOT reward extra detail beyond the expected scope; faithfulness to the expected output's brevity and scope is part of correctness. Minor paraphrasing and stylistic differences are acceptable.
  \end{tabular}}
  \caption{Evaluation instructions provided to the LLM judge for G-Eval. The prompt guides the model to assess semantic agreement between the predicted and reference design intents while penalizing verbosity and mismatched intent.}
  \label{fig:app:g-eval-prompt}
\end{figure}

\begin{figure*}[t!]
  \centering
  \includegraphics[width=\linewidth]{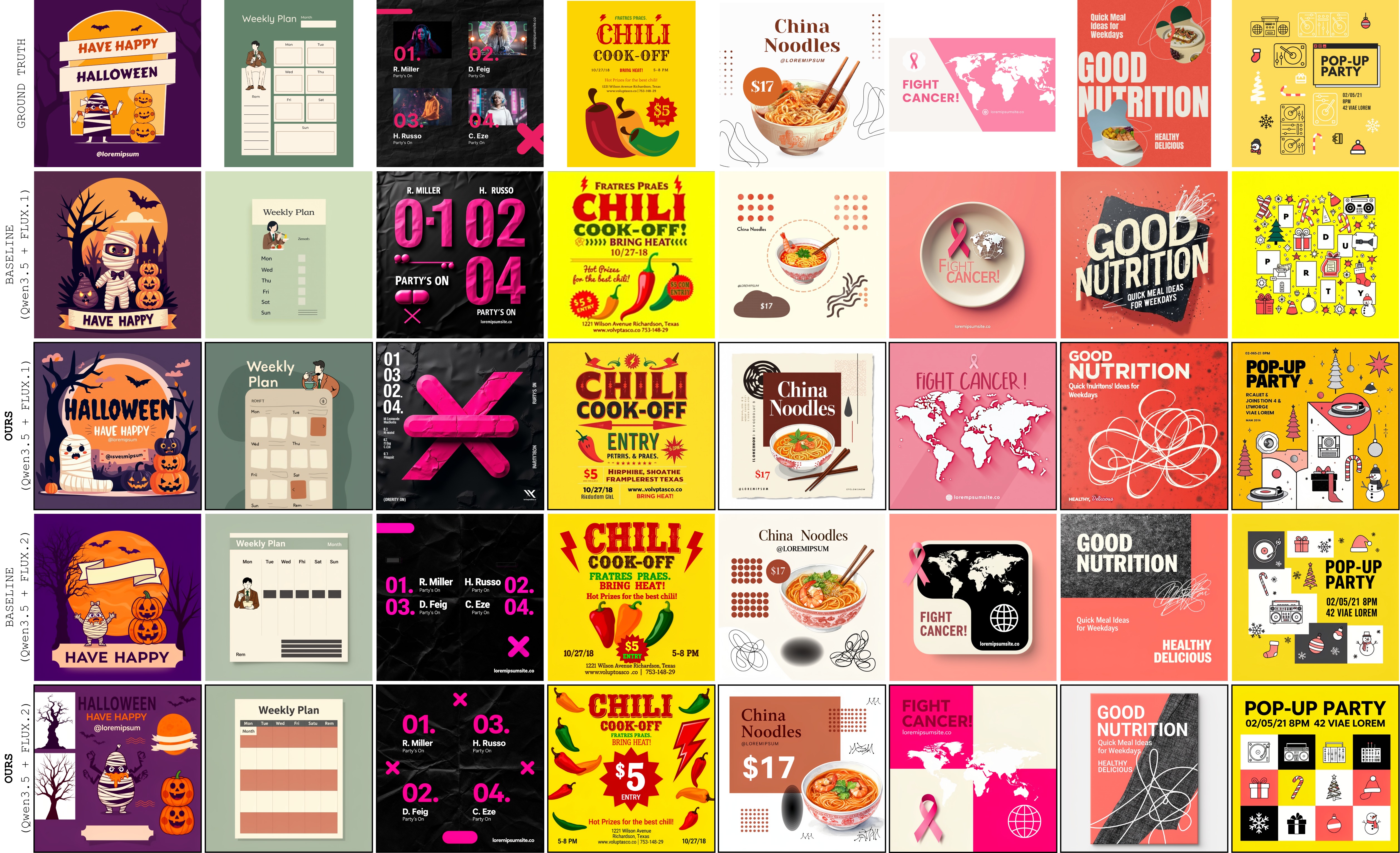}
  \caption{\textbf{Additional qualitative results.} We compare the ground-truth reference design, the baseline prompt policy with FLUX.1-dev, our prompt policy optimized with \textsc{GDB-Reward} using the same generator, the baseline prompt policy with FLUX.2-dev, and our optimized prompt policy with FLUX.2-dev.}
  \label{fig:app:qualitative}
\end{figure*}

\begin{table}[h]
  \centering
  \small
  \begin{tabular}{c|p{0.8\columnwidth}}
    \toprule
    Score & Expected Outcome \\
    \midrule
    \midrule
    0--2 & Wrong/contradictory intent, or a verbose description/caption that does not read as a concise intent statement. \\
    3--5 & Captures the goal but is markedly more verbose/long than the expected output, or misses a core element. \\
    6--8 & Matches the intent and is reasonably concise; only minor verbosity or a minor missing element. \\
    9--10 & Fully captures the design goal in a single concise statement matching the expected output's brevity and scope. \\
    \bottomrule
  \end{tabular}
  \caption{Scoring rubric used by the G-Eval judge to assess design-intent predictions. Higher scores indicate closer semantic agreement with the reference intent while maintaining comparable brevity.}
  \label{tab:app:g-eval-rubric}
\end{table}

\section{Additional Qualitative Examples}
\label{app:qualitative}

Figure~\ref{fig:app:qualitative} presents additional qualitative comparisons across a diverse range of graphic design tasks, including event posters, planners, promotional flyers, awareness campaigns, and advertisements. We compare the ground-truth reference render against outputs from the baseline prompt policy with FLUX.1-dev, our prompt policy optimized with \textsc{GDB-Reward} using the same generator, the baseline prompt policy with FLUX.2-dev, and our optimized prompt policy with FLUX.2-dev.

A remaining limitation of our approach is that the generator does not receive the original photographic assets when they are part of the design. Instead, the prompt policy only provides textual descriptions of these elements, which can lead to larger deviations from the reference render for image-centric designs. This is particularly evident in examples such as the third column, where the reference contains portraits of individual DJs that cannot be reproduced exactly from text alone.

Despite these limitations, the optimized prompt policy improves alignment with the intended design structure and aesthetics. While the frozen image generator remains a bottleneck for exact reproduction of complex layouts, typography, and embedded imagery, our method effectively steers generation toward outputs that better satisfy the visual constraints captured by the design specification.

\section{Additional Quantitative Results}

\begin{table*}[t]
\centering
\small
\setlength{\tabcolsep}{5pt}
\begin{tabular}{@{}lcccccccccc@{}}
\toprule
 & & \multicolumn{5}{c}{Perceptual quality} &
\multicolumn{2}{c}{Rendering fidelity} & \multicolumn{2}{c}{Spatial accuracy} \\
\cmidrule(lr){3-7}\cmidrule(lr){8-9}\cmidrule(lr){10-11}
Model & Full & DS & LP & SS & IR & NM & $\Delta E_{00}$ & OCR & C$_g$ & C$_m$ \\
\midrule
\midrule
\textsc{GDB-Reward} (FLUX.2)
 & \textbf{0.647} & \textbf{0.565} & \textbf{0.296} & \textbf{0.735} & \textbf{0.571} & 0.415 & \textbf{0.578} & \textbf{0.869} & \textbf{0.731} & \textbf{0.675} \\

\textsc{GDB-Reward} (FLUX.1)
 & 0.559 & 0.513 & 0.272 & 0.730 & 0.545 & 0.427 & 0.504 & 0.643 & 0.601 & 0.612 \\

 \midrule
Base (Qwen3.5-9B + SDXL)
 & 0.334 & 0.307 & 0.235 & 0.691 & 0.322 & \textbf{0.493} & 0.526 & 0.139 & 0.281 & 0.232 \\
$+$\textsc{GDB-Reward}
 & 0.446 & 0.359 & 0.257 & 0.717 & 0.436 & 0.487 & 0.537 & 0.255 & 0.513 & 0.464 \\
\bottomrule
\end{tabular}
\caption{\textbf{Quantitative comparison across image-generation backbones on the held-out test split.}
We evaluate the composite \textsc{GDB-Reward} score (\emph{Full}) and its constituent metrics. The top block reports our method with the FLUX.2-dev and FLUX.1-dev generators; the bottom block reports the SDXL backbone before (\emph{Base}) and after (\emph{$+$\textsc{GDB-Reward}}) prompt optimization, holding the policy LLM, reward, and all GRPO hyperparameters fixed. SDXL underperforms the FLUX backbones and, owing to its 77-token CLIP text encoder, cannot ingest the long descriptive prompts our policy learns to produce.}
\label{tab:app:sdxl}
\end{table*}

\begin{figure}[h!]
  \centering
  \includegraphics[width=\linewidth]{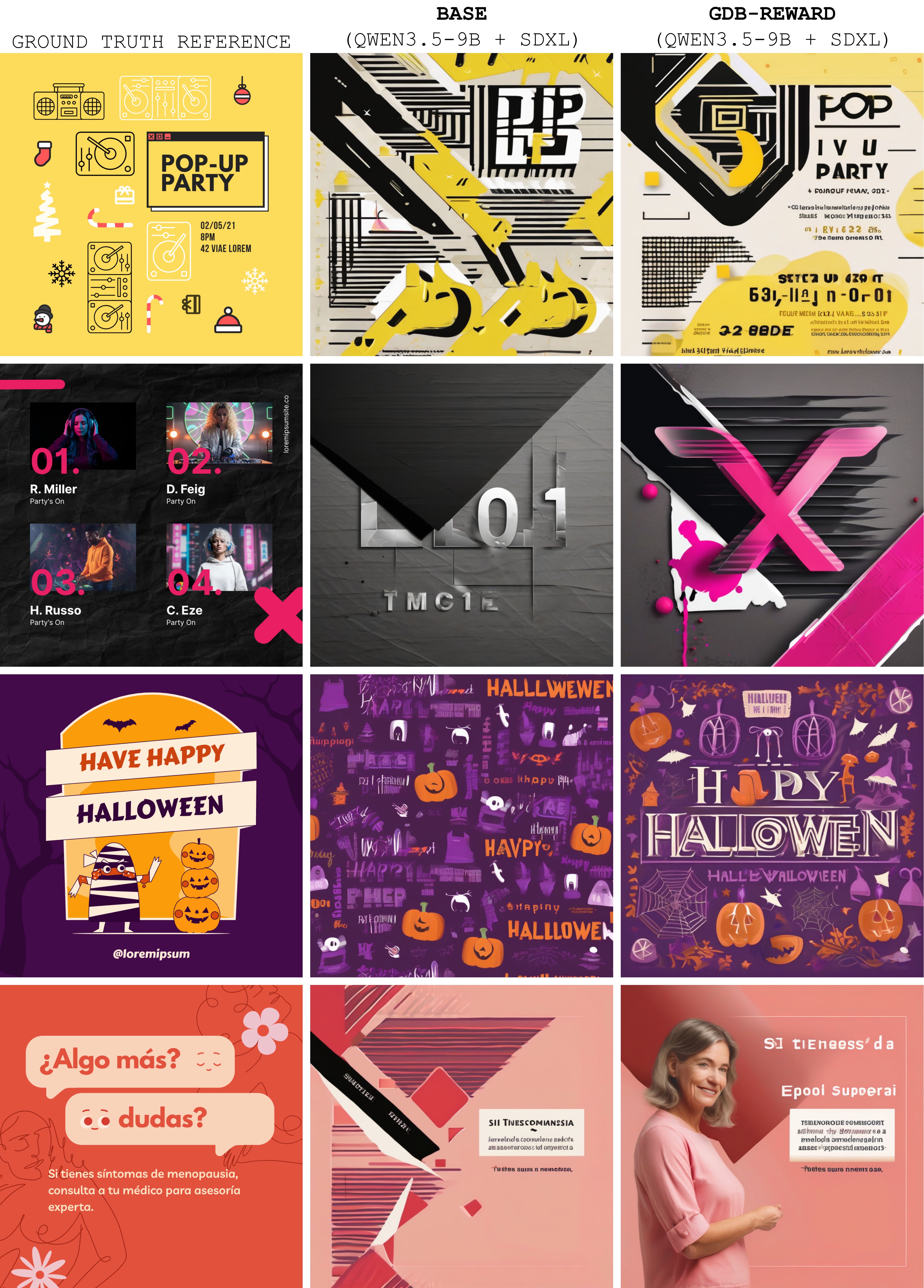}
  \caption{\textbf{Qualitative results on SDXL backbone.} We compare the ground-truth reference design, the baseline prompt policy with SDXL, and our prompt policy optimized with \textsc{GDB-Reward} using the same generator.}
  \label{fig:app:sdxl-qualitative}
\end{figure}

To test whether our GRPO prompt-optimization framework generalizes beyond the FLUX backbone, we repeat the full pipeline with Stable Diffusion XL (stable-diffusion-xl-base-1.0) as the frozen image generator. The setup is deliberately held identical to the FLUX runs to keep the comparison controlled: the same policy LLM (Qwen3.5-9B with an $r{=}16$ LoRA adapter), the same GraphicDesignBench-derived composite reward and its per-concept weighting, and the same GRPO hyperparameters (group size 12, two layouts per step, learning rate $3{\times}10^{-5}$, KL coefficient 0.02), data split, and random seed. Only the image generator is swapped, and its inference settings are adjusted to SDXL-appropriate values rather than copied from FLUX: a classifier-free guidance scale of 5.0 (versus 3.5 for FLUX, which visibly washes SDXL out) and 30 denoising steps (versus 20), with the generator run at $768{\times}768$ to match the FLUX training resolution. We report these results in the supplementary material rather than the main paper for two reasons. First, SDXL's generations are substantially weaker than FLUX's across our metrics, and since FLUX is the current state of the art for text-to-image generation we prioritize it in the main paper. Second, and more fundamentally, SDXL encodes prompts with two CLIP text encoders capped at 77 tokens, whereas FLUX uses a T5 encoder that admits far longer inputs; SDXL therefore cannot ingest the long, richly descriptive prompts our policy learns to produce, meaning it is structurally unable to take full advantage of the prompt-optimization system our method is built around. We include SDXL here only to demonstrate that the method is backbone-agnostic in principle, while noting that a backbone with a short text-encoder context is a poor fit for prompt-level optimization. We provide qualitative results of the SDXL backbone in Figure~\ref{fig:app:sdxl-qualitative}.

\newpage

\end{document}